\documentclass[submission,copyright,creativecommons]{eptcs}

\providecommand{\event}{ACL2 2014} % Name of the event you are submitting to
\providecommand{\urlalt}[2]{\href{#1}{#2}}
\providecommand{\doi}[1]{doi:\urlalt{http://dx.doi.org/#1}{#1}}

\usepackage{acl2}
\usepackage{xspace}
\usepackage{graphicx}

\title{Enhancements to ACL2 in Versions 6.2, 6.3, and 6.4}
\author{
    Matt Kaufmann\\
    \institute{Dept. of Computer Science,\\
               University of Texas at Austin}
    \email{kaufmann@cs.utexas.edu}
    \and
    J Strother Moore\\
    \institute{Dept. of Computer Science,\\
               University of Texas at Austin}
    \email{moore@cs.utexas.edu}
    }
\def\titlerunning{Enhancements to ACL2}
\def\authorrunning{Matt Kaufmann and J Strother Moore}

\begin{document}
\maketitle

\begin{abstract}

% Much shorter abstract than in 2013, but I think that's fine.
We report on improvements to ACL2 made since the 2013 ACL2 Workshop.

\end{abstract}

\section{Introduction}\label{intro}

We discuss ACL2 enhancements introduced in versions released between
the ACL2 Workshops in May, 2013 and July, 2014: Versions 6.2 (June,
2013), 6.3 (October, 2013), and 6.4 (January, 2014).  (These
enhancements were thus made after the release of ACL2 Version 6.1 in
February, 2013.)  Hence this paper is analogous to two papers that
correspond to earlier sets of
releases~\cite{EPTCS114.1,DBLP:journals/corr/abs-1110-4673}.  We
summarize some of the more interesting of the roughly 100 items in the
\href{http://www.cs.utexas.edu/users/moore/acl2/current/manual/index.html?topic=ACL2\_\_\_\_RELEASE-NOTES}{\underline{release
    notes}} for these three releases.  While those release notes
themselves are summaries, they are intended solely as documentation
for what has changed: individual paragraphs are not generally intended
to make sense except to those who have relevant background knowledge.
Moreover, for the sake of completeness those paragraphs are often
rather verbose and dense.  This paper, on the other hand, is intended
to provide a quick way to get up-to-date on the more interesting of
the recent ACL2 system improvements.  For anyone familiar with ACL2,
each brief explanation below is intended to provide sufficient
information so that even inexperienced ACL2 users can get a quick
sense of the enhancement.  Then they can decide whether to read more
in the release notes, or even to follow the provided hyperlinks to
relevant documentation.

%% (defxdoc note-6-2
%% ; Total number of release note items: 41.

%% (defxdoc note-6-3
%% ; Total number of release note items: 36.

%% (defxdoc note-6-4
%% ; Total number of release note items: 27.

ACL2 is typically revised to support soundness, robustness, and
functionality of the system, often in direct response to user
requests.  Because of the maturity of ACL2, the system has many
features, and thus improvements often pertain to aspects of ACL2 that
may be unfamiliar to many ACL2 users, especially novice users.  Our
intention, however, is that this paper will have value to the entire
ACL2 community, including newcomers.  This paper thus includes a few
introductory words about most every feature discussed.  Moreover, the
online version of this paper contains many links to
\href{http://www.cs.utexas.edu/users/moore/acl2/current/manual/index.html?topic=COMMON-LISP\_\_\_\_DOCUMENTATION}{\underline{documentation}}
topics from the online ACL2 User's Manual~\cite{acl2-users-manual},
where one may learn more about those features.  Notice that we use
underlining, as for ``release notes'' and ``documentation'' above, to
mark those hyperlinks.  (The topic names typically match the printed
names, but not always; for example, ``release notes'' above
corresponds to the topic RELEASE-NOTES, which includes a hyphen.)

The focus of this paper is on the user level: our aim is to help ACL2
users to understand what is new with ACL2 (at least to them), in order
to increase its effective use in their applications.  On the other
hand, those who wish to learn about implementation-level changes can
look for Lisp comments in topics note-6-2, note-6-3, and note-6-4 in
Community Book
\href{https://acl2-books.googlecode.com/svn/trunk/system/doc/acl2-doc.lisp}{\underline{\texttt{books/system/doc/acl2-doc.lisp}}}.

One concept that arises several times in this paper is that of a
\href{http://www.cs.utexas.edu/users/moore/acl2/current/manual/index.html?topic=ACL2\_\_\_\_STOBJ}{\underline{{\em stobj}}},
or {\em Single-Threaded OBJect}.  Let us review that concept briefly.
Logically, a stobj is just a list whose members are called its {\em
  fields}.  However, syntactic restrictions in their use permit stobjs
to be modified destructively, and even to have fields that are
represented as (destructively modifiable) arrays in the underlying
host Lisp.

We organize the paper along the lines of the release notes, as
follows.
\begin{itemize}

\item Changes to Existing Features
\item New Features
\item Heuristic Improvements
\item Bug Fixes
\item Changes at the System Level
\item Emacs Support

\end{itemize}

We do not discuss one other category in the release notes,
Experimental/Alternate versions, though there have been improvements
to
\href{http://www.cs.utexas.edu/users/moore/acl2/current/manual/index.html?topic=ACL2\_\_\_\_HONS-AND-MEMOIZATION}{\underline{ACL2(h)}},
\href{http://www.cs.utexas.edu/users/moore/acl2/current/manual/index.html?topic=ACL2\_\_\_\_PARALLELISM}{\underline{ACL2(p)}},
and
\href{http://www.cs.utexas.edu/users/moore/acl2/current/manual/index.html?topic=COMMON-LISP\_\_\_\_REAL}{\underline{ACL2(r)}}.

A label precedes each item being discussed, to indicate the relevant
version of ACL2 in case you wish to read a longer description for that
item in the
\href{http://www.cs.utexas.edu/users/moore/acl2/current/manual/index.html?topic=ACL2\_\_\_\_RELEASE-NOTES}{\underline{release
    notes}}.  For example, if the label is ``6.2'' then you can read
more about the change by visiting the documentation topic,
\href{http://www.cs.utexas.edu/users/moore/acl2/current/manual/index.html?topic=ACL2\_\_\_\_NOTE-6-2}{note-6-2}.

\section{Changes to Existing Features}

As the sets of ACL2 users and projects continue to expand, we learn of
ways to improve its existing capabilities.  Some of these improvements
are rather technical and hence may be new to the reader, who can
follow links below to learn about them.  Our message is twofold:

\begin{itemize}

\item ACL2 offers many capabilities beyond ``only'' a proof engine;
  and

\item if you have avoided features that seemed awkward to use,
  consider trying them again, because they might have improved.

\end{itemize}

\noindent We give a few examples here, referring the reader to the
release notes for a more complete list of such improvements.

\begin{itemize}

\item [6.2] The
  \href{http://www.cs.utexas.edu/users/moore/acl2/current/manual/index.html?topic=ACL2\_\_\_\_TRACE\_42}{\underline{\texttt{trace\$}}}
  utility, which shows calls and return values for indicated
  functions, can be configured to give less noisy output.

\item [6.2] The
  \href{http://www.cs.utexas.edu/users/moore/acl2/current/manual/index.html?topic=ACL2\_\_\_\_GUARD-DEBUG}{\underline{guard-debug}}
  utility, which shows the origin of proof obligations generated for
  \href{http://www.cs.utexas.edu/users/moore/acl2/current/manual/index.html?topic=ACL2\_\_\_\_GUARD}{\underline{guard}}
  verification, avoids duplicating that information.

\item [6.2]
  \href{http://www.cs.utexas.edu/users/moore/acl2/current/manual/index.html?topic=ACL2\_\_\_\_LD}{\underline{\texttt{Ld}}},
  which invokes a read-eval-print-loop, has a new keyword argument,
  \texttt{:ld-missing-input-ok}, which avoids treating a missing file
  as an error.

\item [6.2]
  \href{http://www.cs.utexas.edu/users/moore/acl2/current/manual/index.html?topic=ACL2\_\_\_\_EC-CALL}{\underline{\texttt{Ec-call}}},
  a wrapper for executing function calls in the logic, allows
  non-\href{http://www.cs.utexas.edu/users/moore/acl2/current/manual/index.html?topic=ACL2\_\_\_\_STOBJ}{\underline{stobj}}
  arguments in stobj positions.

\item [6.2] Technical improvements have been made to the
  \href{http://www.cs.utexas.edu/users/moore/acl2/current/manual/index.html?topic=ACL2\_\_\_\_META-EXTRACT}{\underline{\texttt{meta-extract}}}
  capabilities for using facts from the context or world when proving
  meta-level rules (i.e., of class
\href{http://www.cs.utexas.edu/users/moore/acl2/current/manual/index.html?topic=ACL2\_\_\_\_META}{\underline{\texttt{meta}}}
or
\href{http://www.cs.utexas.edu/users/moore/acl2/current/manual/index.html?topic=ACL2\_\_\_\_CLAUSE-PROCESSOR}{\underline{\texttt{clause-processor}}}).

\item [6.3] The
  \href{http://www.cs.utexas.edu/users/moore/acl2/current/manual/index.html?topic=ACL2\_\_\_\_DMR}{\underline{\texttt{dmr}}}
  utility, which supports dynamically watching the rewrite stack, has
  undergone a few improvements.  For example, when the debugger is
  enabled by evaluating \texttt{(dmr-start)}, then subsequent
  evaluation of \texttt{(dmr-stop)} will (once again) disable the
  debugger.

\item [6.3]
  \href{http://www.cs.utexas.edu/users/moore/acl2/current/manual/index.html?topic=ACL2\_\_\_\_BIND-FREE}{\underline{\texttt{Bind-free}}},
  a construct that generates a binding alist for free variables in rule
  hypotheses, can return a list of binding alists to try.

\item [6.3] Evaluation of an event
  \texttt{(\href{http://www.cs.utexas.edu/users/moore/acl2/current/manual/index.html?topic=COMMON-LISP\_\_\_\_PROGN}{\underline{progn}}
    event$_1$ ... event$_k$)} prints each \texttt{event$_i$}
  immediately before evaluating it, just as was already done by
  evaluating a call of
  \href{http://www.cs.utexas.edu/users/moore/acl2/current/manual/index.html?topic=ACL2\_\_\_\_ENCAPSULATE}{\underline{\texttt{encapsulate}}}.

\item [6.3] The
  \href{http://www.cs.utexas.edu/users/moore/acl2/current/manual/index.html?topic=ACL2\_\_\_\_SET-INHIBIT-WARNINGS}{\underline{\texttt{set-inhibit-warnings}}}
  utility, which suppresses specified types of warnings, is
  more predictable.  Moreover, it now has a
  non-\href{http://www.cs.utexas.edu/users/moore/acl2/current/manual/index.html?topic=ACL2\_\_\_\_LOCAL}{\underline{\texttt{local}}}
  variant,
  \href{http://www.cs.utexas.edu/users/moore/acl2/current/manual/index.html?topic=ACL2\_\_\_\_SET-INHIBIT-WARNINGS\_12}{\underline{\texttt{set-inhibit-warnings!}}}.

\item [6.3] Failure messages printed for
  \href{http://www.cs.utexas.edu/users/moore/acl2/current/manual/index.html?topic=COMMON-LISP\_\_\_\_DEFUN}{\underline{\texttt{defun}}}
  indicate which proof attempt fails: termination or
  \href{http://www.cs.utexas.edu/users/moore/acl2/current/manual/index.html?topic=ACL2\_\_\_\_GUARD}{\underline{\texttt{guard}}}
  verification for the indicated definition.

\item [6.3] The functionality of
  \href{http://www.cs.utexas.edu/users/moore/acl2/current/manual/index.html?topic=ACL2\_\_\_\_MAKE-EVENT}{\underline{\texttt{make-event}}},
  a macro-like capability that can involve the ACL2
  \href{http://www.cs.utexas.edu/users/moore/acl2/current/manual/index.html?topic=ACL2\_\_\_\_STATE}{\underline{state}}
  object, has been significantly expanded (see
  Section~\ref{new-features}).

\item [6.4] The
  \texttt{:}\href{http://www.cs.utexas.edu/users/moore/acl2/current/manual/index.html?topic=ACL2\_\_\_\_PBT}{\texttt{\underline{pbt}}}
  (``print back through'') utility, which queries the session
  \href{http://www.cs.utexas.edu/users/moore/acl2/current/manual/index.html?topic=ACL2\_\_\_\_HISTORY}{\underline{history}},
  now abbreviates bodies of large
  \href{http://www.cs.utexas.edu/users/moore/acl2/current/manual/index.html?topic=ACL2\_\_\_\_DEFCONST}{\texttt{\underline{defconst}}}
  forms.

\item [6.4] The utility
  \href{http://www.cs.utexas.edu/users/moore/acl2/current/manual/index.html?topic=ACL2\_\_\_\_SET-INHIBIT-OUTPUT-LST}{\underline{\texttt{set-inhibit-output-lst}}},
  which supports which type of output to suppress, has had the output
  type ``expansion'' replaced by ``history''.

\item [6.4] The optional
  \texttt{:}\href{http://www.cs.utexas.edu/users/moore/acl2/current/manual/index.html?topic=ACL2\_\_\_\_LOOP-STOPPER}{\underline{\texttt{loop-stopper}}} field of a
  \href{http://www.cs.utexas.edu/users/moore/acl2/current/manual/index.html?topic=ACL2\_\_\_\_REWRITE}{\underline{rewrite}}
  rule can specify certain functions to ignore when comparing terms in
  order to avoid looping.  Now, each such ``function symbol'' can
  actually be a
  \href{http://www.cs.utexas.edu/users/moore/acl2/current/manual/index.html?topic=ACL2\_\_\_\_MACRO-ALIASES-TABLE}{\underline{macro
      alias}}.

\item [6.4] ACL2 has a
  \texttt{:}\href{http://www.cs.utexas.edu/users/moore/acl2/current/manual/index.html?topic=ACL2\_\_\_\_LOGIC}{\underline{\texttt{logic}}}
  mode utility,
  \href{http://www.cs.utexas.edu/users/moore/acl2/current/manual/index.html?topic=ACL2\_\_\_\_ORACLE-APPLY}{\underline{\texttt{oracle-apply}}},
  for making higher-order function calls.  This utility now has
  more appropriate restrictions codified in its
\href{http://www.cs.utexas.edu/users/moore/acl2/current/manual/index.html?topic=ACL2\_\_\_\_GUARD}{\underline{guard}}.

\item [6.x] Error handling is improved for several utilities,
including:

\begin{itemize}

\item [6.2] errors from the run-time type-checking utility,
  \href{http://www.cs.utexas.edu/users/moore/acl2/current/manual/index.html?topic=COMMON-LISP\_\_\_\_THE}{\underline{\texttt{THE}}},
  have been eliminated when guard-checking is \texttt{:none};

\item [6.2]
  a much more instructive error message is printed for
  the \href{http://www.cs.utexas.edu/users/moore/acl2/current/manual/index.html?topic=ACL2\_\_\_\_DEFPKG}{\underline{\texttt{defpkg}}}
  ``reincarnation'' error that is encountered upon attempting to
\href{http://www.cs.utexas.edu/users/moore/acl2/current/manual/index.html?topic=ACL2\_\_\_\_PACKAGE-REINCARNATION-IMPORT-RESTRICTIONS}{\underline{redefine a previously-defined package}};

\item [6.2] permission problems no longer cause errors
  for the file utilities
  \href{http://www.cs.utexas.edu/users/moore/acl2/current/manual/index.html?topic=ACL2\_\_\_\_OPEN-INPUT-CHANNEL}{\underline{\texttt{open-input-channel}}}
  and
  \href{http://www.cs.utexas.edu/users/moore/acl2/current/manual/index.html?topic=ACL2\_\_\_\_OPEN-OUTPUT-CHANNEL}{\underline{\texttt{open-output-channel}}}; and

\item [6.4] errors are more instructive when permission problems are encountered
  for
  \href{http://www.cs.utexas.edu/users/moore/acl2/current/manual/index.html?topic=ACL2\_\_\_\_INCLUDE-BOOK}{\underline{\texttt{include-book}}}.

\end{itemize}

\end{itemize}

\section{New Features}\label{new-features}

This section focuses on a few of the more interesting new features
recently added to ACL2.  As before, the release notes for the
indicated versions contain more complete information.

\begin{itemize}

\item [6.2] The
  \href{http://www.cs.utexas.edu/users/moore/acl2/current/manual/index.html?topic=ACL2\_\_\_\_DEFSTOBJ}{\underline{\texttt{defstobj}}}
  event, which introduces singled-threaded objects (see
  Section~\ref{intro}), permits
  \href{http://www.cs.utexas.edu/users/moore/acl2/current/manual/index.html?topic=ACL2\_\_\_\_STOBJ}{\underline{stobj}}s
  to have fields that are themselves stobjs or arrays of stobjs.  In
  the case of these {\em nested stobj} structures, fields are accessed
  using a new construct,
  \href{http://www.cs.utexas.edu/users/moore/acl2/current/manual/index.html?topic=ACL2\_\_\_\_STOBJ-LET}{\underline{\texttt{stobj-let}}}.

\item [6.2] ACL2 supports
  \href{http://www.cs.utexas.edu/users/moore/acl2/current/manual/index.html?topic=ACL2\_\_\_\_META}{\underline{meta}}theoretic
  reasoning, which may be implemented using
  \href{http://www.cs.utexas.edu/users/moore/acl2/current/manual/index.html?topic=ACL2\_\_\_\_EXTENDED-METAFUNCTIONS}{\underline{extended
      metafunctions}} that can access the environment.  The logical
  \href{http://www.cs.utexas.edu/users/moore/acl2/current/manual/index.html?topic=ACL2\_\_\_\_WORLD}{\underline{world}}
  --- that is, the ACL2 database --- is now available to such
  functions using the function \texttt{mfc-world}.

\item [6.2] ACL2 supports many kinds of
  \href{http://www.cs.utexas.edu/users/moore/acl2/current/manual/index.html?topic=COMMON-LISP\_\_\_\_DECLARE}{\underline{\texttt{declare}}}
  forms, including not only some that are processed by the host Common
  Lisp compiler, but also others processed by ACL2 that are specified
  using
  \href{http://www.cs.utexas.edu/users/moore/acl2/current/manual/index.html?topic=ACL2\_\_\_\_XARGS}{\underline{\texttt{xargs}}}
  forms.  A new \texttt{xargs} keyword, \texttt{:split-types}, can be
  used to specify that the function's
  \href{http://www.cs.utexas.edu/users/moore/acl2/current/manual/index.html?topic=COMMON-LISP\_\_\_\_TYPE}{\underline{\texttt{type}}}
  declarations should be provable from its
  \href{http://www.cs.utexas.edu/users/moore/acl2/current/manual/index.html?topic=ACL2\_\_\_\_GUARD}{\underline{guard}},
  not add to its guard.

\item [6.2] See
  \href{http://www.cs.utexas.edu/users/moore/acl2/current/manual/index.html?topic=ACL2\_\_\_\_QUICK-AND-DIRTY-SUBSUMPTION-REPLACEMENT-STEP}{\underline{quick-and-dirty-subsumption-replacement-step}}
  for a way to turn off a potentially expensive prover heuristic.

\item [6.3] The macro-like utility
  \href{http://www.cs.utexas.edu/users/moore/acl2/current/manual/index.html?topic=ACL2\_\_\_\_MAKE-EVENT}{\underline{\texttt{make-event}}}
  evaluates a form to obtain a new form --- its {\em expansion} --- and
  then typically submits that expansion.  This utility has been made
  more flexible by the addition of the following new capabilities,
  which we discuss below for the benefit of those who already
  have some familiarity with
  \href{http://www.cs.utexas.edu/users/moore/acl2/current/manual/index.html?topic=ACL2\_\_\_\_MAKE-EVENT}{\underline{\texttt{make-event}}}.

\begin{itemize}

\item Expansions may have the form \texttt{(:or event$_1$
  ... event$_k$)}.  In this case, each \texttt{event$_i$} is evaluated
  in turn until one succeeds.  That \texttt{event$_i$} is then treated
  as the actual expansion, and is not re-evaluated after expansion (as
  would be necessary without the use of \texttt{:or}).

\item Expansions may have the form \texttt{(:do-proofs event)}, for
  insisting on performing proofs during evaluation of \texttt{event}
  even in contexts where proofs are normally skipped (such as when
  \href{http://www.cs.utexas.edu/users/moore/acl2/current/manual/index.html?topic=ACL2\_\_\_\_REBUILD}{\underline{\texttt{rebuild}}}
  is invoked).

\item A keyword argument, \texttt{:expansion?}, provides an
  optimization that can eliminate storing expansions in
  \href{http://www.cs.utexas.edu/users/moore/acl2/current/manual/index.html?topic=ACL2\_\_\_\_CERTIFICATE}{\underline{certificate}}
  files.

\end{itemize}

\item [6.3] See
  \href{http://www.cs.utexas.edu/users/moore/acl2/current/manual/index.html?topic=ACL2\_\_\_\_GET-INTERNAL-TIME}{\underline{get-internal-time}}
  for how to change ACL2's timing reports, so that instead of being
  based on run time (cpu time) they are based on real time (wall-clock
  time).

\item [6.3] A new utility,
  \href{http://www.cs.utexas.edu/users/moore/acl2/current/manual/index.html?topic=ACL2\_\_\_\_SYS-CALL+}{\underline{\texttt{sys-call+}}},
  can invoke a shell command just as is done by
  \href{http://www.cs.utexas.edu/users/moore/acl2/current/manual/index.html?topic=ACL2\_\_\_\_SYS-CALL}{\underline{\texttt{sys-call}}}.
  However, while {\texttt{sys-call}} prints the command's output as a
  side effect but returns \texttt{nil}, {\texttt{sys-call+}} actually
  returns that output as a string.

\item [6.3] A new utility,
  \href{http://www.cs.utexas.edu/users/moore/acl2/current/manual/index.html?topic=ACL2\_\_\_\_VERIFY-GUARDS\_B2}{\underline{\texttt{verify-guards+}}},
  is like the
  \href{http://www.cs.utexas.edu/users/moore/acl2/current/manual/index.html?topic=ACL2\_\_\_\_VERIFY-GUARDS}{\underline{\texttt{verify-guards}}}
  utility for verifying that calls of the indicated function lead only
  to function calls that satisfy their
  \href{http://www.cs.utexas.edu/users/moore/acl2/current/manual/index.html?topic=ACL2\_\_\_\_GUARD}{\underline{guard}}s
  (preconditions).  However, in \texttt{verify-guards+}, that argument
  can be the
  \href{http://www.cs.utexas.edu/users/moore/acl2/current/manual/index.html?topic=ACL2\_\_\_\_MACRO-ALIASES-TABLE}{\underline{macro
      alias}} for a function.  See
  \href{http://www.cs.utexas.edu/users/moore/acl2/current/manual/index.html?topic=ACL2\_\_\_\_VERIFY-GUARDS}{\underline{\texttt{verify-guards}}}
  for an example showing why it would be unsound to permit
  \texttt{verify-guards} to take a macro alias as its argument.

\item [6.4] The
  \href{http://www.cs.utexas.edu/users/moore/acl2/current/manual/index.html?topic=ACL2\_\_\_\_BOOKDATA}{\underline{bookdata}}
  utility writes out event data on a per-book basis, for tools such as
  the one written by Dave Greve that is found in directory
  \texttt{tools/book-conflicts/} of the
  \href{http://www.cs.utexas.edu/users/moore/acl2/current/manual/index.html?topic=ACL2\_\_\_\_COMMUNITY-BOOKS}{\underline{Community
      Books}}.

\item [6.4] The utilities
  \href{http://www.cs.utexas.edu/users/moore/acl2/current/manual/index.html?topic=ACL2\_\_\_\_ADD-INCLUDE-BOOK-DIR}{\underline{\texttt{add-include-book-dir}}}
  and
  \href{http://www.cs.utexas.edu/users/moore/acl2/current/manual/index.html?topic=ACL2\_\_\_\_DELETE-INCLUDE-BOOK-DIR}{\underline{\texttt{delete-include-book-dir}}},
  whose effects are
  \href{http://www.cs.utexas.edu/users/moore/acl2/current/manual/index.html?topic=ACL2\_\_\_\_LOCAL}{\underline{local}}
  to a book, specify directories denoted by \texttt{:dir} arguments of
  \href{http://www.cs.utexas.edu/users/moore/acl2/current/manual/index.html?topic=ACL2\_\_\_\_INCLUDE-BOOK}{\underline{\texttt{include-book}}}.
  These utilities now have
  non-\href{http://www.cs.utexas.edu/users/moore/acl2/current/manual/index.html?topic=ACL2\_\_\_\_LOCAL}{\underline{local}}
  analogues,
  \href{http://www.cs.utexas.edu/users/moore/acl2/current/manual/index.html?topic=ACL2\_\_\_\_ADD-INCLUDE-BOOK-DIR\_12}{\underline{\texttt{add-include-book-dir!}}}
  and
  \href{http://www.cs.utexas.edu/users/moore/acl2/current/manual/index.html?topic=ACL2\_\_\_\_DELETE-INCLUDE-BOOK-DIR\_12}{\underline{\texttt{delete-include-book-dir!}}}.

\item [6.4]
  \href{http://www.cs.utexas.edu/users/moore/acl2/current/manual/index.html?topic=ACL2\_\_\_\_CONGRUENCE}{\underline{Congruence}}
  rules specify
  \href{http://www.cs.utexas.edu/users/moore/acl2/current/manual/index.html?topic=ACL2\_\_\_\_EQUIVALENCE}{\underline{equivalence}}
  relations to maintain during rewriting.~\cite{congruence}
  The functionality of congruence rules is extended by 
  \href{http://www.cs.utexas.edu/users/moore/acl2/current/manual/index.html?topic=ACL2\_\_\_\_PATTERNED-CONGRUENCE}{\underline{patterned
      congruence}} rules~\cite{patterned-congruence-rules} 
  allowing a more general
  specification of where an
  \href{http://www.cs.utexas.edu/users/moore/acl2/current/manual/index.html?topic=ACL2\_\_\_\_EQUIVALENCE}{\underline{equivalence}}
  is to be maintained.
% --- for example, when rewriting \texttt{y} in a
%  term \texttt{(mv-nth 1 (f (cons u v) y 'a))}.

\end{itemize}

\section{Heuristic Improvements}\label{heuristic}

ACL2 development began in 1989, and development for the Boyer-Moore
series of provers began in 1971.  It is therefore not surprising, at
least to us, that there are relatively few changes to prover
heuristics in recent ACL2 releases.  However, there are a few, because
the user community finds new applications of ACL2 that present
opportunities for improving the heuristics.  The following summary is
intended to give a sense of how ACL2's heuristics have been tweaked,
without diving into unnecessary details of how they work.  (Heuristics
are generally not documented at the user level, since we do not expect
it to be necessary or useful to understand in any depth how they work
in order to be an effective ACL2 user.)

\begin{itemize}

\item [6.2] ACL2 has an {\em ancestors check} heuristic that can
  prevent excessive backchaining through hypotheses of rules.  The
  following list (which we invite beginners to skip!) describes ways
  in which this heuristic has been improved:

  \begin{itemize}

  \item the heuristic no longer allows failure for
  \href{http://www.cs.utexas.edu/users/moore/acl2/current/manual/index.html?topic=ACL2\_\_\_\_FORCE}{\underline{\texttt{force}}}d
  hypotheses;

  \item it is delayed until a quick
  (\href{http://www.cs.utexas.edu/users/moore/acl2/current/manual/index.html?topic=ACL2\_\_\_\_TYPE-SET}{\underline{type-set}})
  check has a chance to verify or refute the hypothesis; and

  \item a slight weakening of the heuristic now permits backchaining
    based on counting variable occurrences.

  \end{itemize}

\item [6.2] The context
  (\href{http://www.cs.utexas.edu/users/moore/acl2/current/manual/index.html?topic=ACL2\_\_\_\_TYPE-ALIST}{\underline{type-alist}})
  built from a goal now considers assumptions in a different order
  that may strengthen that context.

\item [6.3] \texttt{:By}
  \href{http://www.cs.utexas.edu/users/moore/acl2/current/manual/index.html?topic=ACL2\_\_\_\_HINTS}{\underline{hints}}
  are intended to specify when a goal is subsumed by a known fact,
  such as the formula of a
  \href{http://www.cs.utexas.edu/users/moore/acl2/current/manual/index.html?topic=ACL2\_\_\_\_DEFTHM}{\underline{\texttt{defthm}}}
  event.  The subsumption check for \texttt{:by} hints has been made
  less restrictive.

\item [6.3] The hint \texttt{:do-not preprocess} is intended to cause
  the ACL2 prover to skip the {\em preprocess} step of its proof
  \href{http://www.cs.utexas.edu/users/moore/acl2/current/manual/index.html?topic=ACL2\_\_\_\_HINTS-AND-THE-WATERFALL}{\underline{\em
      waterfall}}.  This hint now also eliminates the preprocess step
  during the application of \texttt{:use} and \texttt{:by}
  \href{http://www.cs.utexas.edu/users/moore/acl2/current/manual/index.html?topic=ACL2\_\_\_\_HINTS}{\underline{hints}}.

\end{itemize}

\section{Bug Fixes}

Among the bugs fixed were soundness bugs, which are all mentioned below.  These
tend to be obscure, and most are not generally encountered by users; but we
consider them particularly important to fix.  Among the other bugs
fixed, we mention here only a few of the more interesting ones.

Thus we begin with a description of the soundness bugs that were
fixed.  In contrast to the rest of this paper, we do not attempt to
explain much about these bugs, not even how they could lead to proofs
of \texttt{nil}.  Pointers to some of those proofs can be found in the
release notes.

\begin{itemize}

\item [6.2] System functions \texttt{acl2-magic-mfc} and
  \texttt{acl2-magic-canonical-pathname}, which were not designed for
  direct invocation by the user, could be exploited to prove
  \texttt{nil}.

\item [6.2]
  \href{http://www.cs.utexas.edu/users/moore/acl2/current/manual/index.html?topic=ACL2\_\_\_\_STOBJ}{\underline{Stobj}}s
  could be confused with strings in raw Lisp.

\item [6.3] A
  \href{http://www.cs.utexas.edu/users/moore/acl2/current/manual/index.html?topic=ACL2\_\_\_\_STOBJ}{\underline{stobj}}
  could be bound by a
  \href{http://www.cs.utexas.edu/users/moore/acl2/current/manual/index.html?topic=COMMON-LISP\_\_\_\_LET}{\underline{\texttt{let}}}
  or
  \href{http://www.cs.utexas.edu/users/moore/acl2/current/manual/index.html?topic=ACL2\_\_\_\_MV-LET}{\underline{\texttt{mv-let}}}
  form without being among the outputs of that form, which allowed for
  serious violations of single-threadednesss.

\item [6.3] (Gnu Common Lisp only) There was a bug in Common Lisp code
  in the implementation of the utility,
  \href{http://www.cs.utexas.edu/users/moore/acl2/current/manual/index.html?topic=ACL2\_\_\_\_SET-DEBUGGER-ENABLE}{\underline{\texttt{set-debugger-enable}}}.

\item [6.4] ACL2 supports rules of class
  \texttt{:}\href{http://www.cs.utexas.edu/users/moore/acl2/current/manual/index.html?topic=ACL2\_\_\_\_DEFINITION}{\underline{\texttt{definition}}},
  which are typically equalities that are much like normal
  definitions, but can actually have hypotheses.  A soundness bug
  resulted from incorrect application of such rules during the
  handling of \texttt{:expand}
  \href{http://www.cs.utexas.edu/users/moore/acl2/current/manual/index.html?topic=ACL2\_\_\_\_HINTS}{\underline{hints}}.

\item [6.4] A
  \href{http://www.cs.utexas.edu/users/moore/acl2/current/manual/index.html?topic=ACL2\_\_\_\_STOBJ}{\underline{stobj}}
  recognizer predicate could be violated after updating the stobj.

\item [6.4] The checks made when admitting
  \href{http://www.cs.utexas.edu/users/moore/acl2/current/manual/index.html?topic=ACL2\_\_\_\_CONGRUENCE}{\underline{congruence}}
  rules failed to ensure that a certain variable on the right-hand
  side of the conclusion, which was intended to be a fresh variable,
  was indeed actually fresh (i.e., did not occur elsewhere in the
  rule).

\end{itemize}

\noindent Here are a few of the more interesting bugs that were fixed, other
than soundness bugs.

\begin{itemize}

\item [6.2] ACL2 supports a notion of {\em abstract}
  \href{http://www.cs.utexas.edu/users/moore/acl2/current/manual/index.html?topic=ACL2\_\_\_\_STOBJ}{\underline{stobj}},
  which is an abstraction of a corresponding ordinary ({\em concrete})
  stobj.  The
  \href{http://www.cs.utexas.edu/users/moore/acl2/current/manual/index.html?topic=ACL2\_\_\_\_DEFABSSTOBJ}{\underline{\texttt{defabbstobj}}}
  event, which introduces a new abstract stobj, incurs certain proof
  obligations to ensure proper correspondence between the new abstract
  stobj and its specified concrete stobj.  These proof obligations, in
  the form of
  \href{http://www.cs.utexas.edu/users/moore/acl2/current/manual/index.html?topic=ACL2\_\_\_\_DEFTHM}{\underline{\texttt{defthm}}}
  \href{http://www.cs.utexas.edu/users/moore/acl2/current/manual/index.html?topic=ACL2\_\_\_\_EVENTS}{\underline{events}},
  are printed when submitting a \texttt{defabbstobj} event, except for
  events that have themselves already been admitted.  However, the
  events printed were not always sufficient in order to admit the
  \texttt{defabbstobj} event.

\item [6.3] ACL2's proof output indicates
  \href{http://www.cs.utexas.edu/users/moore/acl2/current/manual/index.html?topic=ACL2\_\_\_\_SPLITTER}{\underline{splitter}}s:
  rule applications that generate more than one subgoal.  However,
  splitters were sometimes alleged when only a single subgoal was
  generated.

\item [6.3] The utility
  \href{http://www.cs.utexas.edu/users/moore/acl2/current/manual/index.html?topic=ACL2\_\_\_\_WOF}{\underline{\texttt{wof}}},
  for directing output to a file (as the name stands for ``With Output
  to File''), could cause an error when no error was appropriate.
  This problem also occurred with the
  \texttt{:}\href{http://www.cs.utexas.edu/users/moore/acl2/current/manual/index.html?topic=ACL2\_\_\_\_PSOF}{\underline{\texttt{psof}}}
  (``Print Saved Output to File'')
  utility, since
  \texttt{psof} is defined in terms of \texttt{wof}.
  (\texttt{:Psof} is similar to the utility
  \texttt{:}\href{http://www.cs.utexas.edu/users/moore/acl2/current/manual/index.html?topic=ACL2\_\_\_\_PSO}{\underline{\texttt{pso}}}
  (``Print Saved Output''), as both print proof output that had been
  suppressed by
  \href{http://www.cs.utexas.edu/users/moore/acl2/current/manual/index.html?topic=ACL2\_\_\_\_GAG-MODE}{\underline{gag-mode}};
  but \texttt{:psof} prints saved proof output to a file, while
  \texttt{:pso} prints to the terminal, which can take much longer.)

\end{itemize}

\section{Changes at the System Level}\label{system}

Some of the more sweeping changes to ACL2 have taken place far from
its theorem prover.  Here we touch briefly on just a few of those.

\begin{itemize}

\item [6.2] From the beginning of ACL2, Gnu Common Lisp (GCL) has been
  among the host Lisps on which ACL2 can be built.  Now, the ANSI
  version of GCL is also a host Lisp on which ACL2 (and ACL2(h)) can
  be built.

\item [6.2] The previous system for certifying the
  \href{http://www.cs.utexas.edu/users/moore/acl2/current/manual/index.html?topic=ACL2\_\_\_\_COMMUNITY-BOOKS}{\underline{Community
      Books}} has been updated; in particular, it is largely based on
  \texttt{cert.pl} and other utilities maintained by Jared Davis and
  Sol Swords.  See
  \href{http://www.cs.utexas.edu/users/moore/acl2/current/manual/index.html?topic=ACL2\_\_\_\_BOOKS-CERTIFICATION}{\underline{books-certification}}.

\item [6.3] ACL2 is now available between releases, via SVN, at
  \href{http://acl2-devel.googlecode.com}{\underline{\texttt{http://acl2-devel.googlecode.com}}}.
  Disclaimer (as per the warning message printed at startup): {\em The
    authors of ACL2 consider svn distributions to be experimental;
    they may be incomplete, fragile, and unable to pass our own
    regression.}  That said, we have seen few problems with SVN
  distributions.

\item [6.4] The ACL2 documentation is now maintained in the XDOC
  format developed and implemented by Jared Davis.  Indeed, it is now
  recommended to peruse the
  \href{http://www.cs.utexas.edu/users/moore/acl2/current/combined-manual/index.html}{\underline{acl2+books
      combined manual}}.  That manual includes not only the ACL2
  User's Manual but also topics from the
  \href{http://www.cs.utexas.edu/users/moore/acl2/current/manual/index.html?topic=ACL2\_\_\_\_COMMUNITY-BOOKS}{\underline{Community
      Books}}, such as the
  \href{http://www.cs.utexas.edu/users/moore/acl2/current/combined-manual/index.html}{\underline{XDOC}}
  topic itself as well as
  \href{http://www.cs.utexas.edu/users/moore/acl2/current/combined-manual/index.html?topic=ACL2\_\_\_\_CERT.PL}{\underline{\texttt{cert.pl}}}
  (mentioned above).  Note that many books now include an XDOC book,
  which causes the \texttt{:doc} command to be an alias for the
  \texttt{:xdoc} command using the
  \href{http://www.cs.utexas.edu/users/moore/acl2/current/combined-manual/index.html?topic=ACL2\_\_\_\_LD-KEYWORD-ALIASES}{\underline{\texttt{ld-keyword-aliases}}}
  feature.  This causes a bit of noisy output upon first invocation.

\end{itemize}

\section{Emacs Support}

The primary Emacs-related ACL2 enhancement is a new utility introduced
in Version 6.4,
\href{http://www.cs.utexas.edu/users/moore/acl2/current/manual/index.html?topic=ACL2\_\_\_\_ACL2-DOC}{\underline{ACL2-Doc}},
for browsing documentation inside Emacs.  This utility takes the
place of Emacs Info, which is no longer supported because of the
transition to XDOC discussed in Section~\ref{system}.  Emacs users
will find this utility to be a nice alternative to using
\texttt{:}\href{http://www.cs.utexas.edu/users/moore/acl2/current/manual/index.html?topic=ACL2\_\_\_\_DOC}{\underline{\texttt{doc}}}
at the terminal.  It can be used to browse either the
\href{http://www.cs.utexas.edu/users/moore/acl2/current/combined-manual/index.html}{\underline{acl2+books
    combined manual}} or the
\href{http://www.cs.utexas.edu/users/moore/acl2/current/manual/index.html}{\underline{ACL2
    User's Manual}}.  This utility is loaded automatically into Emacs
when loading the standard file
\href{https://acl2-devel.googlecode.com/svn/trunk/emacs/emacs-acl2.el}{\underline{\texttt{emacs/emacs-acl2.el}}}.

\section{Conclusion}

We have outlined some of the more interesting and important changes to
ACL2 in Versions 6.2, 6.3, and 6.4.  We hope that a quick read of this
paper will enable ACL2 users to focus quickly on topics of particular
interest, while easily following links (in the online version of this
paper) in order to learn more about those topics, with the result of
becoming more effective ACL2 users.  Many more changes (about 100
altogether) are described in the
\href{http://www.cs.utexas.edu/users/moore/acl2/current/manual/index.html?topic=ACL2\_\_\_\_RELEASE-NOTES}{\underline{release
    notes}} for these versions, and many changes at a lower level are
described in comments in the source code for those release notes, such
as \texttt{(defxdoc note-6-2 ...)}, in Community Book
\href{https://acl2-books.googlecode.com/svn/trunk/system/doc/acl2-doc.lisp}{\underline{\texttt{books/system/doc/acl2-doc.lisp}}}.

A critical component in the continued evolution of ACL2 is feedback
from its user community, which we hope will continue!  We also
appreciate contributions by the user community to the large body of
Community Books~\cite{acl2-books-svn}.  These books put demands on the system
and help us to test improvements.

\section*{Acknowledgements}

% Wording follows previous paper (but names have changed), which is
% fine with me and J.

We thank members of the ACL2 community whose feedback has led us to
continue making improvements to ACL2.  In particular, we thank the
following, who are the people mentioned in one or more specific items
in the release notes for Version 6.2, 6.3, or 6.4: Harsh Raju
Chamarthi, Jared Davis, Jen Davis, Caleb Eggensperger, Shilpi Goel,
Dave Greve, Warren Hunt, Robert Krug, Camm Maguire, David Rager,
Gisela Rossi, Sol Swords, Raymond Toy, and Nathan Wetzler,

We expressly thank Warren Hunt for his continuing support of ACL2 use and
development for many years at UT Austin.

We are grateful to Shilpi Goel, Warren Hunt, Robert Krug, Sandip Ray,
Nathan Wetzler, and the referees for feedback on drafts of this paper.

This material is based upon work supported by DARPA under Contract
No. N66001-10-2-4087, by ForrestHunt, Inc., and by the National
Science Foundation under Grant No. CCF-1153558.

%\section{Bibliography}

% The line
% \nocite{*}
% was in the example file.  However, quoting
% http://www.math.uiuc.edu/~hildebr/tex/bibliographies0.html:
%  The command \nocite{*} causes all items in the database to be
%  included in the references, regardless of whether or not they are
%  cited in the paper.
% We don't want that!
\bibliographystyle{eptcs}
\bibliography{kaufmann-moore}
\end{document}